\documentclass[11pt,a4paper]{article}
\usepackage[utf8]{inputenc}
\usepackage[english]{babel}
\usepackage{amsfonts}
\usepackage{authblk}
\usepackage{fullpage}
\usepackage{graphicx}   
\usepackage{tabularx} 
\usepackage{subfigure}
\usepackage{multirow}
\usepackage{wrapfig}   
\usepackage{hyperref} 
\usepackage{tikz,pgfplots}

\usepackage{algorithmic,algorithm}
\usepackage{amsmath} 
\usepackage{amssymb}  

\newcommand{\eg}{\textit{e.g. }}
\newcommand{\ie}{\textit{i.e. }}  
\newcommand{\comment}[1]{}  

\title{Expected exponential loss for gaze-based video and volume ground truth annotation}
\author[1]{Laurent Lejeune}
\author[2]{Mario Christoudias}
\author[1]{Raphael Sznitman}
\affil[1]{University of Bern, Switzerland}
\affil[2]{Weather Analytics, USA}

\date{}

\begin{document}
\maketitle
\begin{abstract}
Many recent machine learning approaches used in medical imaging are highly reliant on large amounts of image  and groundtruth data. In the context of object segmentation, pixel-wise annotations are extremely expensive to collect, especially in video and 3D volumes. To reduce this annotation burden, we propose a novel framework to allow annotators to simply observe the object to segment and record where they have looked at with a \$200 eye gaze tracker. Our method then estimates pixel-wise probabilities for the presence of the object throughout the sequence from which we train a classifier in semi-supervised setting using a novel Expected Exponential loss function. We show that our framework provides superior performances on a wide range of medical image settings compared to existing strategies and that our method can be combined with current crowd-sourcing paradigms as well.
\end{abstract}

\section{Introduction}  
\label{sec:intro} 

\begin{figure}[t!]
\centering
\includegraphics[width=0.24\textwidth]{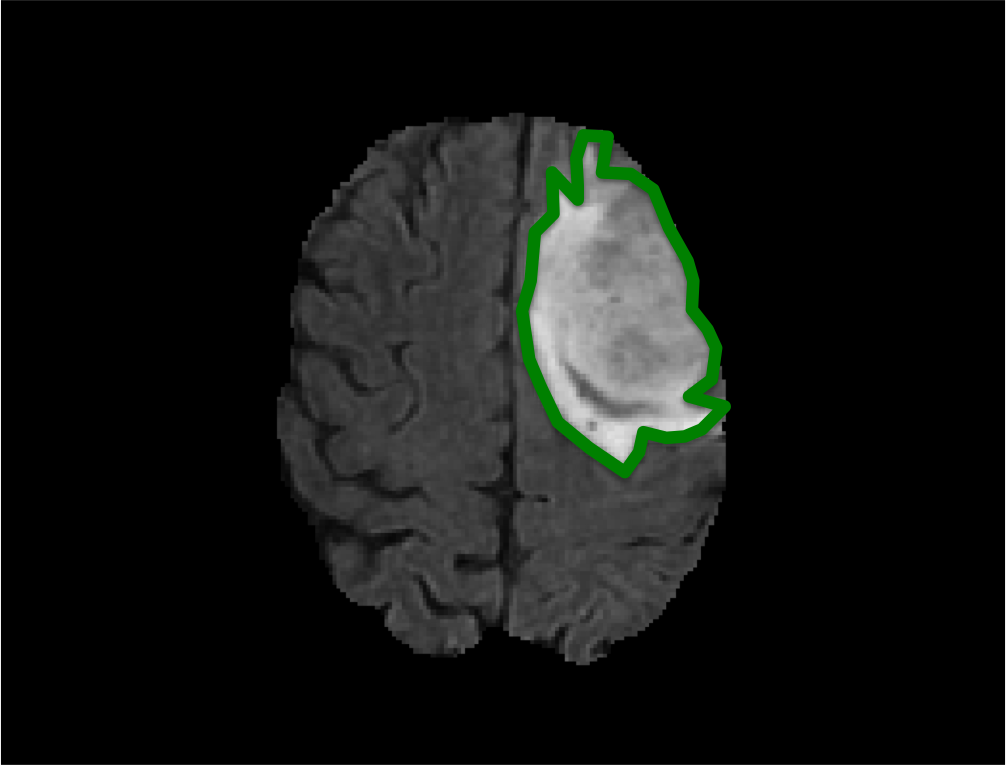}
\includegraphics[width=0.24\textwidth]{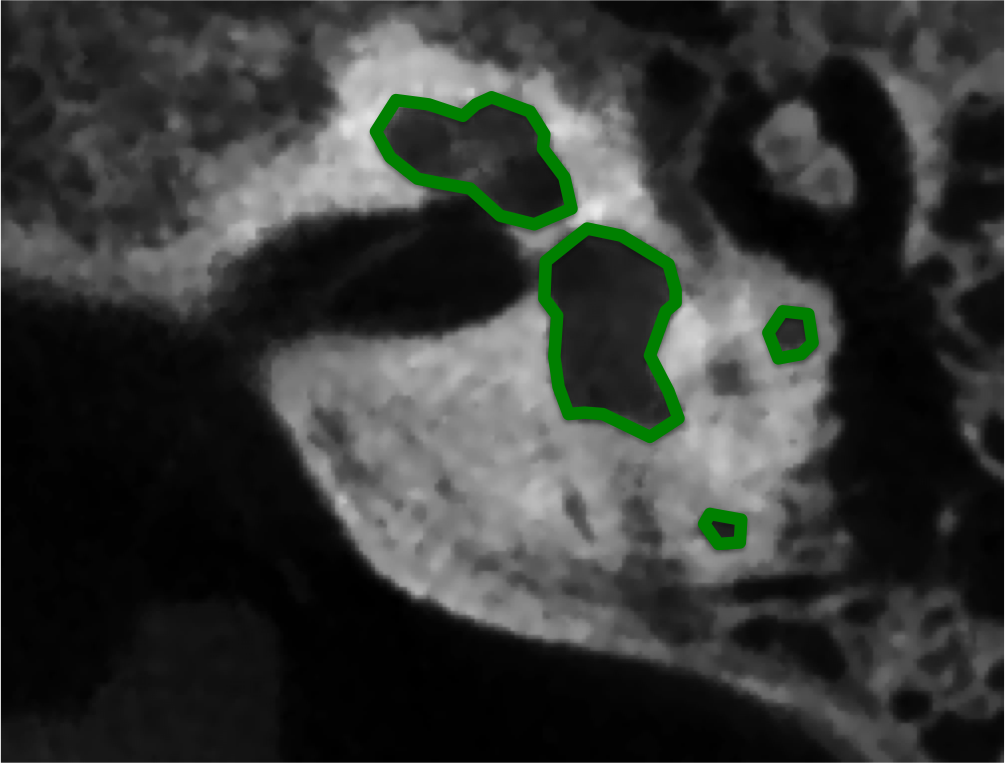}
\includegraphics[width=0.24\textwidth]{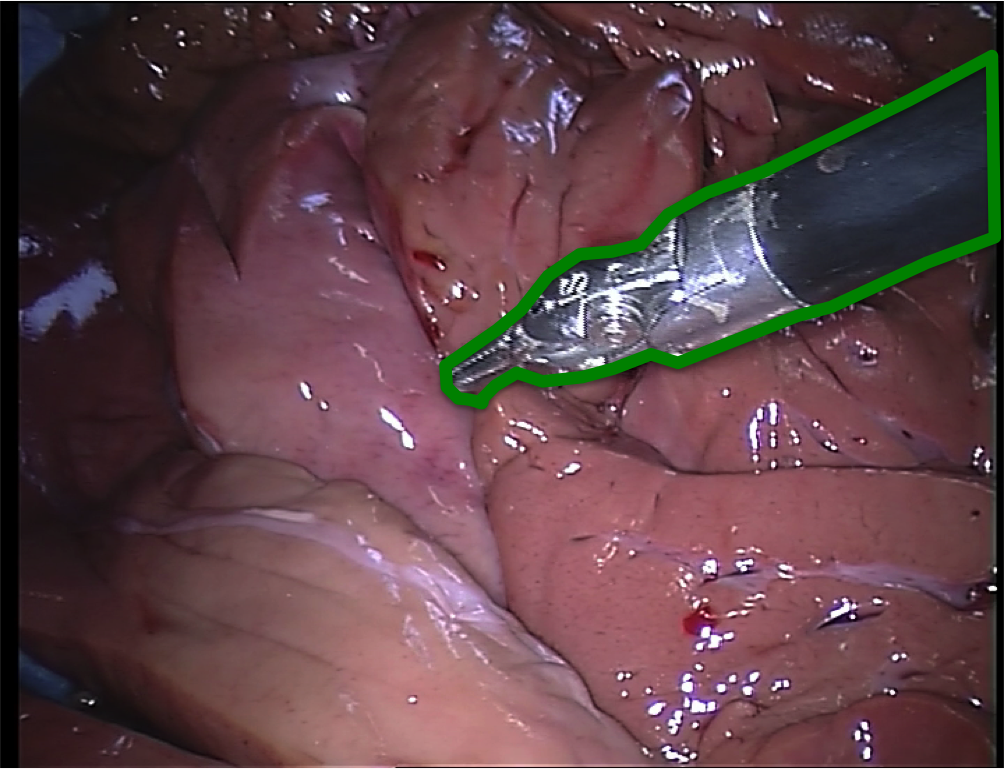}
\includegraphics[width=0.24\textwidth]{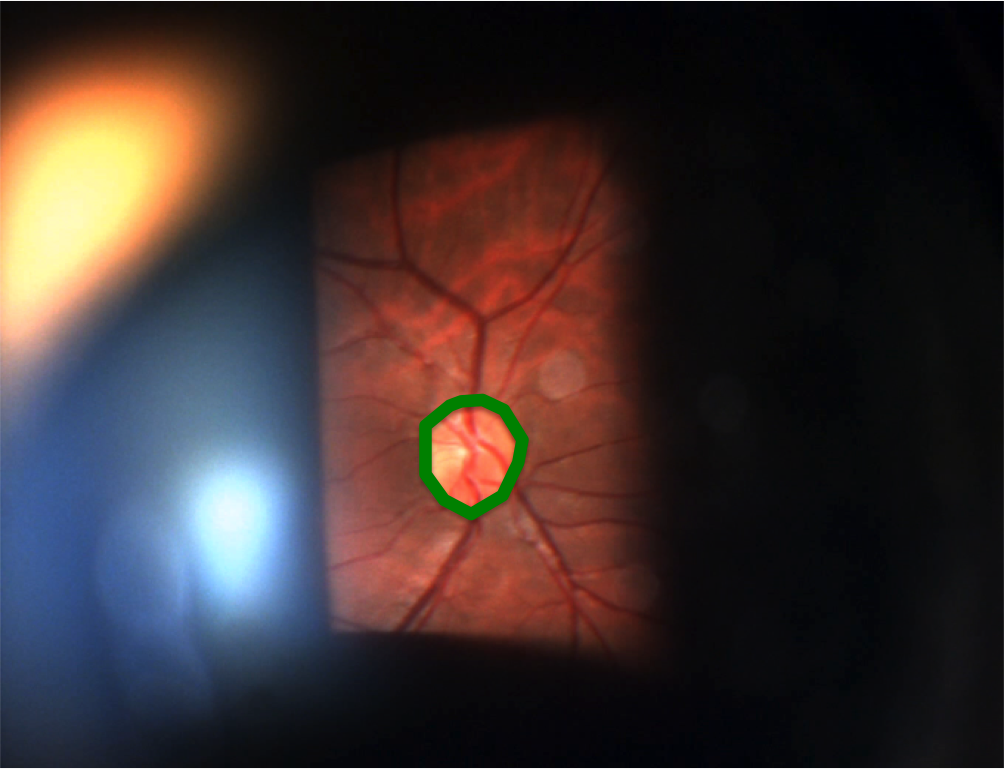}
\caption{Examples of volume and video data with a structure to annotate. Left to right: Brain tumor (3D-MRI), cochlea (3D-CT); surgical instrument (video endoscopy); optic disc (video microscope). Green contours highlight ideal ground truth regions.}
\label{fig:examples}
\end{figure}

Ground truth annotations play a critical role in the development of machine learning methods in medical imaging. Indeed, advances in deep learning strategies, coupled with the advent of image data in medicine have greatly improved performances for tasks such as structure detection and anatomical segmentation across most imaging modalities (\eg MRI, CT, Endoscopy, Microscopy)~\cite{kamnitsas17,anthimopoulos16}. Yet the process of acquiring ground truth data or annotations remains laborious and challenging, especially in video and 3D image data such as those depicted in Fig.~\ref{fig:examples}.

To mitigate manual annotation dependence, semi- and unsupervised methods have been key research areas to reduce the overall annotation burden placed on domain experts (\eg radiologist, biologist, surgeon etc....). Most notably, Active Learning (AL)~\cite{KonSznFua15,MosSzn16}, Transfer Learning (TL)~\cite{ShiRoth16,BerBecSalFua2016} and Crowd-Sourcing (CS)~\cite{MaierHein2014,Cheplygina2016} provide frameworks for learning with either limited or noisy ground truth data and have been applied to a larger number of applications. Yet in AL domain experts are still necessary to actively provide ground truth data, sequentially or in batch. Similarly, CS relies on manual annotators to follow carefully crafted labeling tasks in order to leverage non-experts, which produces highly variable ground truth quality~\cite{Cheplygina2016}. 

Alternatively, Vilari{\~n}o et al.~\cite{FerVil07} used an eye gaze tracker to annotate polyps in video colonoscopy. In their approach an expert passively viewed a video and stared at polyps. From these, they trained an SVM classifier to label the sequence, treating regions around each gaze location as positives and the rest of the image domain as negative samples. As we show in our experiments however, this approach is limited to detecting objects of fixed size and does not extend well to pixel-wise segmentation tasks. Also related is the work in~\cite{sadegh09} which mapped out regions of interest using a gaze tracker on individual frames observed for extended periods of time. More recently, the work of~\cite{khosravan16} is closely related to our setting, with the important distinction that our data is viewed in one pass and applied to video and volumetric data. 

To overcome this limitation, we propose a novel framework to produce pixel-wise segmentation for an object present in a volume (or video sequence) using gaze observations collected from a \$200 off-the-shelf gaze tracker. Assuming a single target is present throughout the image data, we cast our problem as a semi-supervised problem where samples are either labeled as positive (gazed image locations) or unknown (the rest of the image data which could be positive or negative). To learn in this regime, we introduce a new Expected Exponential loss function that can be used within a traditional gradient boosting framework. In particular, the expectation is taken with respect to the unknown labels, requiring a label probability estimate. We describe how to estimate these with a novel strategy and show that our approach not only provides superior performances over existing methods in a variety of medical imaging modalities (\ie laparoscopy, microscopy, CT and MRI) but can be used in a crowd-sourcing context as well.

\section{Gaze-based pixel-wise annotation} 
\label{sec:approach} 
Our goal is to produce a pixel-wise segmentation of a specific structure of interest located in a video or in a volume (\ie we treat a volume as a video sequence). To do this, we ask a domain expert to watch and follow the structure throughout the sequence. While viewing the sequence, we track the persons eye gaze by means of a commercially available eye gaze tracker. In our setting, this provides a single gaze location for each viewed image and we assume the observer is compliant in the task. 

To produce ground truth annotations, we cast this problem as a binary semi-supervised machine learning problem, where one must determine a pixel-wise segmentation of the structure of interest in each of the images using only the sequence itself and the gaze locations. We assume that gaze locations correspond to the structure and propose a novel Expected Exponential loss function that explicitly takes into account that some labels are known while others are not. This loss leverages probability estimates regarding the unknown labels and we present a strategy to estimate these effectively. Note that, we do not focus on learning a function that generalizes to other similar sequences, but one that annotates the given sequence as well as possible.

\begin{figure}[t!]
\centering
\includegraphics[width=0.99\textwidth]{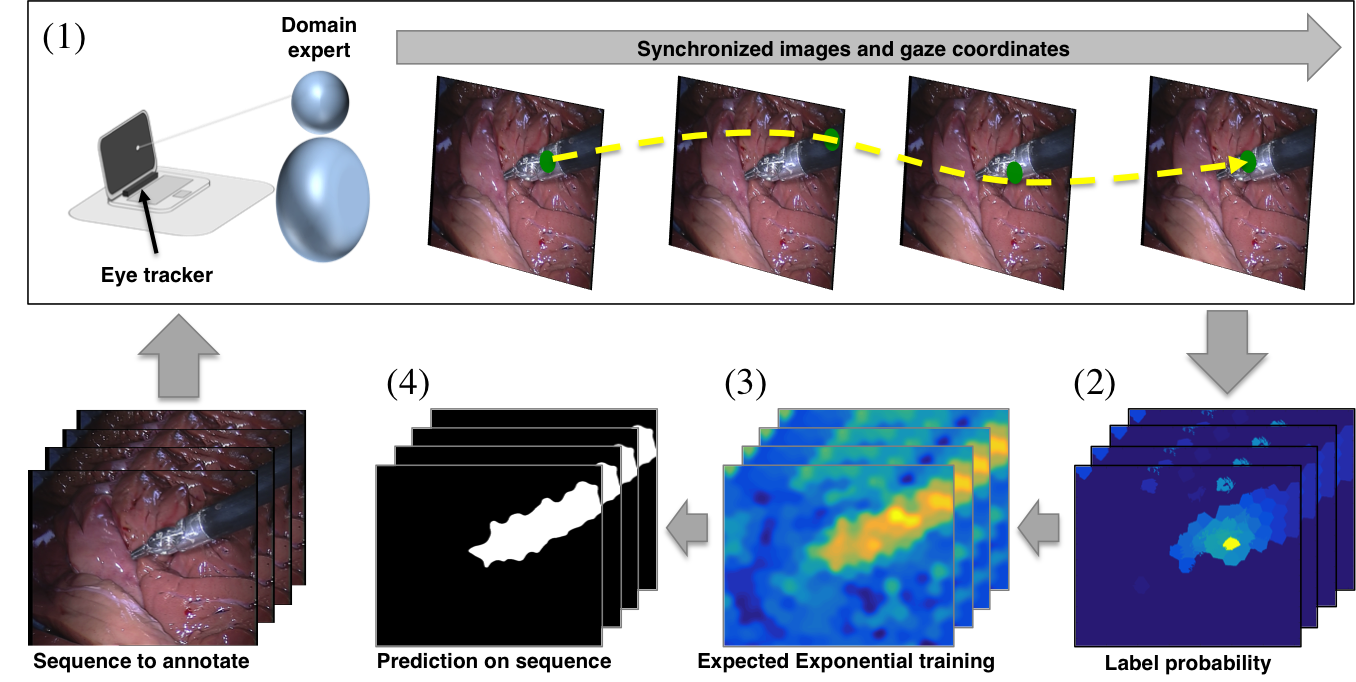}
\caption{Example Gaze-based annotating. A surgical instrument must be annotated in a sequence. (1) A domain expert watches the sequences and has their gaze collected during the viewing. (2) Our method then estimates object likelihoods over the sequence and (3) we then train a classifier with an Expected Exponential loss function using a subset of the image data. (4) The classifier is then used to evaluate the remainder of the sequence.}
\label{fig:concept}
\end{figure}

For a given image sequence, our approach is organized as follows and is illustrated in Fig.~\ref{fig:concept}: (1) the expert views the sequence and 2D gaze locations are collected; (2) we estimate the label probability by using the image data and the gaze information; (3) we then train a gradient boosted classifier with our proposed loss on a subset of the image data; (4) using the trained classifier, we predict the remainder of the image data. We will detail these steps in the following sections but first define some notation used throughout this paper. 

{\bf{Notation}}: Let an image sequence (or volume) be denoted $\mathcal{I} = [I_0,\ldots,I_T]$ and let $\mathcal{G} = \{g_t\}_{t=0}^T$ such that $g_t\in\mathbb{R}^2$ is a 2D gaze pixel location in $I_t$. While we ideally would like a pixel-wise segmentation, we choose to decompose each image using a temporal superpixel strategy~\cite{chang13tsp} and operate at a superpixel level instead. We thus let $I_t$ be described by the set of non-overlapping superpixels $S_t=\{S^n_t\}_{n=0}^{N_t}$ and define the set of all superpixels across all images as $\mathcal{S}=\{S_t\}_{t=0}^T$. We denote the set $\mathcal{P} = \{S^n_t | g_t \subset S^n_t, t=0,\ldots,T,n=0,\ldots N_t \}$ as all superpixels observed and the rest as $\mathcal{U} = \mathcal{S} \setminus \mathcal{P}$. We associate with each $S^n_t$ a binary random variable $Y^n_t\in\{-1,1\}=\mathcal{Y}$, such that $Y^n_t=1$ if $S^n_t$ is part of the object and -1 otherwise. In particular, we defined $Y_n$ as a Bernoulli random variable, $Y^n_t \sim  Ber(\epsilon_{S_t^n}), \epsilon_{S_t^n} \in (0,1)$. Note that for superpixels observed by gaze $S^n_t\in\mathcal{P}$, we consider these as part of the object and let $Y^n_t= 1$ with $\epsilon_{S_t^n}=1$. 

\section{Learning with an Expected Exponential loss}
{\bf{Expected Exponential Loss (EEL)}}:
Our goal is to train a prediction function, $f:\mathcal{S} \rightarrow \mathcal{Y}$ that takes into account observed superpixels as well as the unobserved ones. To do this we propose the following EEL function
\begin{align}
 \mathcal{EE} &= \mathbb{E}^{{Y}} \left[ \sum_{S\in \{\mathcal{P},\mathcal{U}\}} e^{-f(S){{Y}}} \right]
\end{align}
\noindent
where the expectation is taken with respect to all $Y$s. By linearity of expectation and the fact that labeled superpixels have no uncertainty in their label, we can rewrite the loss for all superpixels as
\begin{align}
 \mathcal{EE} &= \sum_{S\in \mathcal{P}} e^{-f(S)} + \sum_{S\in \mathcal{U}} \left( \epsilon_{S} e^{-f(S) Y} + (1-\epsilon_{S})e^{f(S)  Y} \right)
 \label{eq:loss}
\end{align}
\noindent
Note that this Eq.~\eqref{eq:loss} is a generalization of the Exponential Loss (EL)~\cite{elemStat}. In the case where labels are known, the loss is the same as the traditional loss as the expectation is superfluous. For unknown samples, the value of $\epsilon_{S}$ weighs the impact of the superpixels. For instance, if $\epsilon_{S}$ is close to 0.5 then the sample does not affect the loss. Conversely values of $\epsilon_{S}$ close to 1 (or 0) will strongly impact the loss. 

{\bf{Implementation}}: We implemented the above EEL within a traditional Gradient Boosting classifier~\cite{elemStat}, by regressing to the residual given by the derivate of Eq.~\eqref{eq:loss}. For all experiments, we used stumps as weak learners, a shrinkage factor of 1 and the line search was replaced by a constant weight of 1. The weak learner stumps operate on features extracted from the center of the superpixel. In particular, we used generic Overfeat features~\cite{SerEigZha14} which provide a rich characterization of a region and its context (\eg 4086 sized feature vector).

During training we used all superpixels in $\mathcal{P}$ and used 10\% of those in $\mathcal{U}$. A total of 50 boosting rounds was performed in all cases. To predict segmentations for the entire volume, we predicted the remaining 90\% of superpixels in $\mathcal{U}$. 

\section{Probability estimation for unknown labels}
\label{sec:probapropa}
To estimate $\epsilon_{S}$ in Eq.~\eqref{eq:loss}, we take inspiration from the Label Propagation method~\cite{zhouBous03}, which uses a limited number of positive and negative samples to iteratively propagate labels to unobserved samples. In our setting however, we only propagate positive samples to unlabeled samples using the gaze information as well as pixel motion estimation to constrain the probability diffusion.

We let $P_0 = [p_0,\ldots,p_N ] $ be a vector of initial probabilities for all superpixels in a given image, where $p_n = P(Y=1|\mathcal{P})$ is the probability that superpixel $S_n$ is part of the object. In practice, we estimate $p_n$ by computing a gaze dependent Lab color model using all gaze locations and assessing how likely a superpixel $S_n$ is part of the object. That is, we compute
\begin{equation}
p_n = \max_{S \in \mathcal{P} } \mathcal{N}(S_n | \mu_S , \Sigma_S),
\end{equation}
\noindent
where $\mathcal{N}$ is a Gaussian distribution such that $\mu_S$ and $\Sigma_S$ are the color mean and covariance of pixels in a superpixel $S$ that was gazed at. For superpixels that were gazed at, their probability is fixed at 1. To propagate probabilities, we also define a $N \times N$ affinity matrix, denoted $W$ with values
\begin{equation}
w_{ij}=  \exp(-|\theta_i-\theta_j|_2 / 2\sigma_a^2) \exp(-|C(S_i)-C(S_j)|_2  / 2\sigma^2_d),
\end{equation}
\noindent
where for superpixel $S_i$, $C(S_i)$ is its center and $\theta_i$ is its average gradient orientation. In cases where $S_i$ and $S_j$ are separated by more than $\tau$ pixels, $w_{ij}=0$. $\sigma_a$ and $\sigma_d$ are model parameters reflecting the variance in angle difference and the impact of neighboring superpixels, respectively.

\begin{figure}[t!]
\centering
\includegraphics[width=0.98\textwidth]{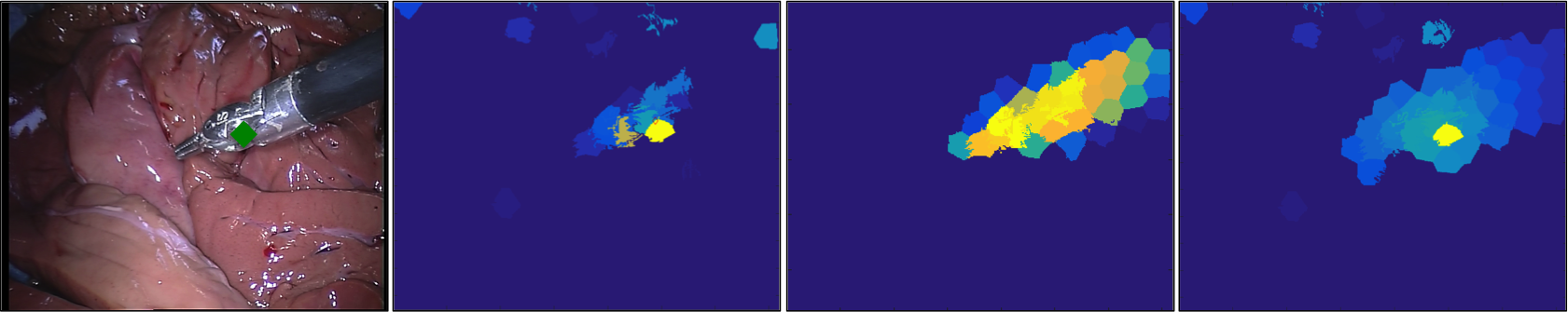}
\caption{Probability propagation. Left to right: original image with the gaze location highlighted in green; Initial $P_0$ estimate from the gaze-based color model; Image regions with high optical flow; Estimated probability after propagation. Dark blue regions depict low probability while warmer regions correspond to higher probabilities.}
\label{fig:probaprop}
\end{figure}

Propagation can then be computed iteratively by solving 
\begin{equation}
P_{m+1} = \alpha \Omega P_m + (1-\alpha)P_0,
\end{equation}
\noindent 
where $\alpha \in (0,1)$ is a diffusion parameter, $D$ is a diagonal matrix with entries $d_{ii} = \sum_{j} w_{ij}$ and $\Omega = D^{-1/2}WD^{-1/2}$. Fig.~\ref{fig:probaprop} shows the initial $P_0$, the associated optical flow regions and the final propagated probability for a given image. While the original method described in~\cite{zhouBous03} hinged on a minimum of one positive and one negative sample to prove the existence of a closed form convergence solution, the same cannot be said of the current setting where no negative samples are known. For this reason, we iterate a total of 10 times and then use the estimates for the $\epsilon_S$ values in Eq.~\eqref{eq:loss}. This value was experimentally determined and shown to perform well for a number of image sequences (see Sec.~\ref{sec:exp}). The process is repeated for all frames in the sequence.

Note that the probability estimate is computed from a single gaze sequence and the corresponding image data. As such, if more than one domain expert viewed the same sequence, as it is the case in Crowd-Sourcing tasks, this process can be repeated for each observer and averaged over all observers. In our experiments, we show that doing so brings increased performances over that given by a single observer.

\section{Experiments} 
\label{sec:exp} 
To evaluate the performance of our method we compare it to the method presented in Vilari{\~n}o et al.~\cite{FerVil07}. We also show how the EEL approach compares with that of using $\epsilon_S$ estimates only (see~Sec.~\ref{sec:probapropa}), as well using $\epsilon_S$, 
with a traditional EL when binarizing the labels using a fixed threshold $\epsilon_S = 0.5$. The following parameters were kept constant: $\alpha = 0.95,\sigma_a = 0.5, \sigma_d = 50, \tau =50$ and the superpixel size was set to match 1$^{\circ}$ on the viewing monitor.

We evaluated each of the above mentioned methods on 4 very different image sequences (see Fig.~\ref{fig:examples} for examples): (1) a 3D brain MRI containing a tumor to annotate from the BRATS challenge~\cite{menze15} consisting of 73 slices, (2) a 30 frame surgical video sequence from the MICCAI EndoVision challenge~\footnote{Endoscopic vision challenge: \url{https://endovis.grand-challenge.org}} where a surgical instrument must be annotated, (3) a 95-slice 3D CT scan where a cochlea must be annotated and (4) a slit-lamp video recording (195 frames) of a human retina where the optic disk must be segmented. Pixel-wise annotated ground truth on all frames of each sequence was either available or produced by a domain expert. In all sequences, one and only one object is present throughout the sequence. 

Our method was implemented in MATLAB and takes roughly 30 minutes to segment a 30 frame volume with 720 $\times$ 576 sized frames, of which the bulk of time is used to compute temporal superpixels and training our classifier. Even though real-time requirements are not necessary in this application, we believe this computation time could be reduced with an improved implementation
\begin{table}[b!]
\centering
\begin{tabular}{ llcccc} 
Dataset~~~~ & Metric~~~ & Vilarino et al.~\cite{FerVil07}~~& Probability Est.~~~~& ~~~~ EL~~~~~~~~&  EEL\\
\hline
\hline
\multicolumn{1}{ l  }{\multirow{2}{*}{Brain tumor} } &
\multicolumn{1}{ l  }{AUC} & 0.687 & 0.963 & 0.974 & \bf{0.976}\\ 
\multicolumn{1}{ c  }{}                        &
\multicolumn{1}{ l }{F-score} & 0.551 & 0.428 & 0.482 & \bf{0.592} \\ 
\hline
\multicolumn{1}{ l  }{\multirow{2}{*}{Cochlea} } &
\multicolumn{1}{ l  }{AUC} & 0.687 & 0.963 & 0.974 & \bf{0.976} \\ 
\multicolumn{1}{ c  }{}                        &
\multicolumn{1}{ l }{F-score} & 0.223 & 0.239 & 0.431 & \bf{0.631}    \\ 
\hline
\multicolumn{1}{ l }{\multirow{2}{*}{Surgical instr.} } &
\multicolumn{1}{ l }{AUC} & 0.346 & 0.949 & 0.959 & \bf{0.985} \\ 
\multicolumn{1}{ c }{}                        &
\multicolumn{1}{ l }{F-score} & 0.239 & 0.711 & 0.725 & \bf{0.851} \\ 
\hline
\multicolumn{1}{ l  }{\multirow{2}{*}{Optic disc} } &
\multicolumn{1}{ l  }{AUC} & 0.687 & 0.963 & 0.974 & \bf{0.976} \\ 
\multicolumn{1}{ c  }{}                        &
\multicolumn{1}{ l }{F-score} & 0.506 & 0.367 & 0.494 & \bf{0.665}  \\ 
\hline
\end{tabular}
\caption{Area Under the Curve (AUC) and  performances for each approach on each dataset. Highlight maximum values in bold.}
\label{tab:stats}
\end{table} 

Gaze locations were collected with an Eye Tribe Tracker (Copenhagen, Denmark) which provide 1$^{\circ}$ degree tracking accuracy. To collect gaze locations, a computer monitor and the tracker was placed roughly 1 meter from the experts face. Device-specific calibration was performed before all recordings (\ie a 2-minute long procedure done once before each viewing). 2D gaze locations were collected and mapped to the viewed image content using the manufacturers API. Domain experts were instructed to stare at the target and avoid looking at non-object image regions. Once each sequence was observed, the different methods inferred the object throughout the entire image data. 

\begin{figure}[b!]
\centering
\includegraphics[width=0.99\textwidth]{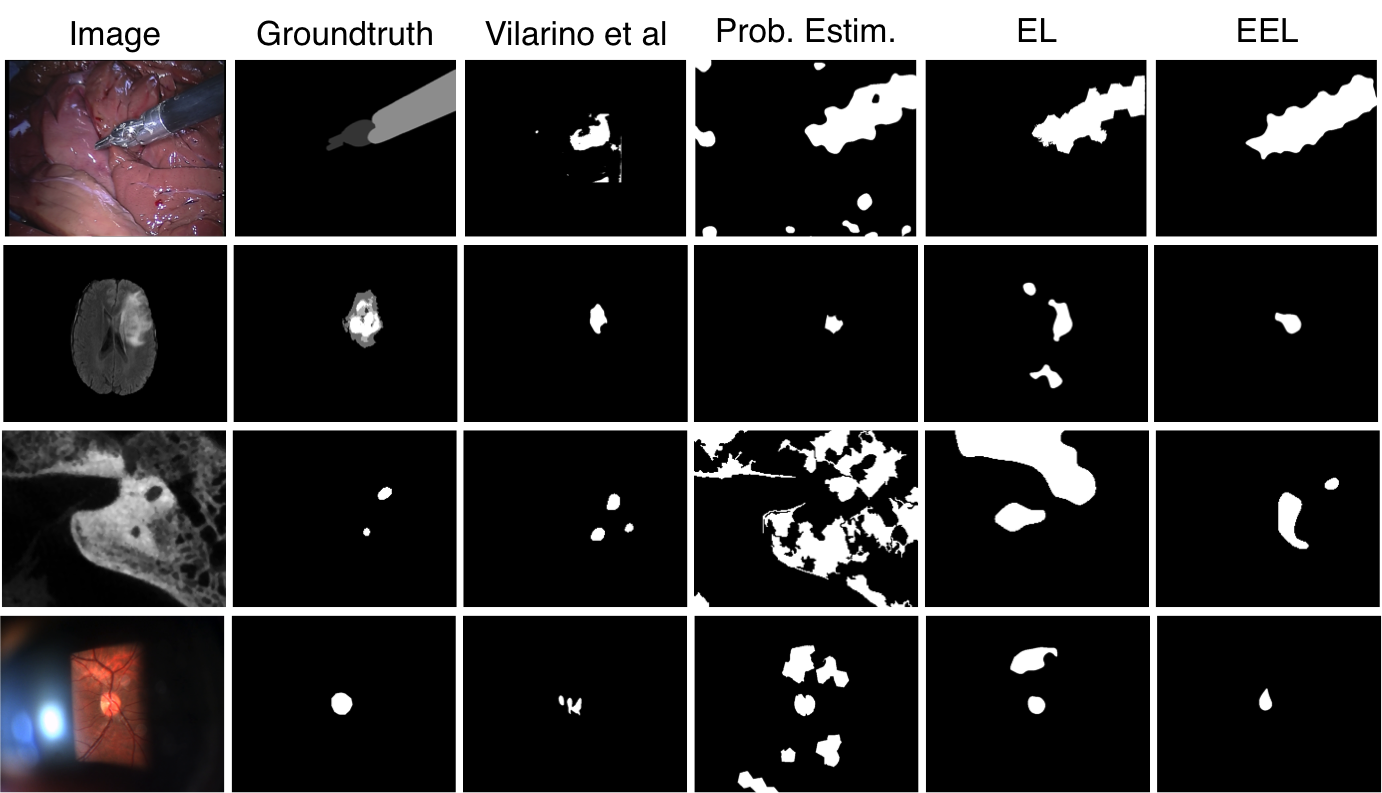}
\caption{Qualitative results. Each row shows a different dataset with an example image, the associated desired ground truth and the produced outcome of~\cite{FerVil07}, using the probability estimation approach, EL approach and the proposed EEL. Binary images were generated by thresholding results at a 5\% False Positive Rate.}
\label{fig:qualresult}
\end{figure}

{\bf{Results -- Annotation accuracy}}: Table~\ref{tab:stats} reports the Area Under the Curve (AUC) and the F-score performances of each method applied to each dataset. In general, we report that the proposed combined label estimation and EEL function provide the highest AUC and F-score values across the tested sequences. Fig.~\ref{fig:qualresult} visually depicts example frames from each sequence and the outcome of each method, as well as the ground truth. To generate these binary images, a \%5 false positive threshold was applied (\ie threshold was determined using the ground truth). One can see that in cases where the object to segment occupies large areas of the image, as is the case for the surgical instrument, both the traditional loss approach and that of~\cite{FerVil07} do not perform as well since they treat significant portions of the background as positive samples during their respective learning phases.  

\begin{figure}[t!]
\centering
\includegraphics[width=0.45\textwidth]{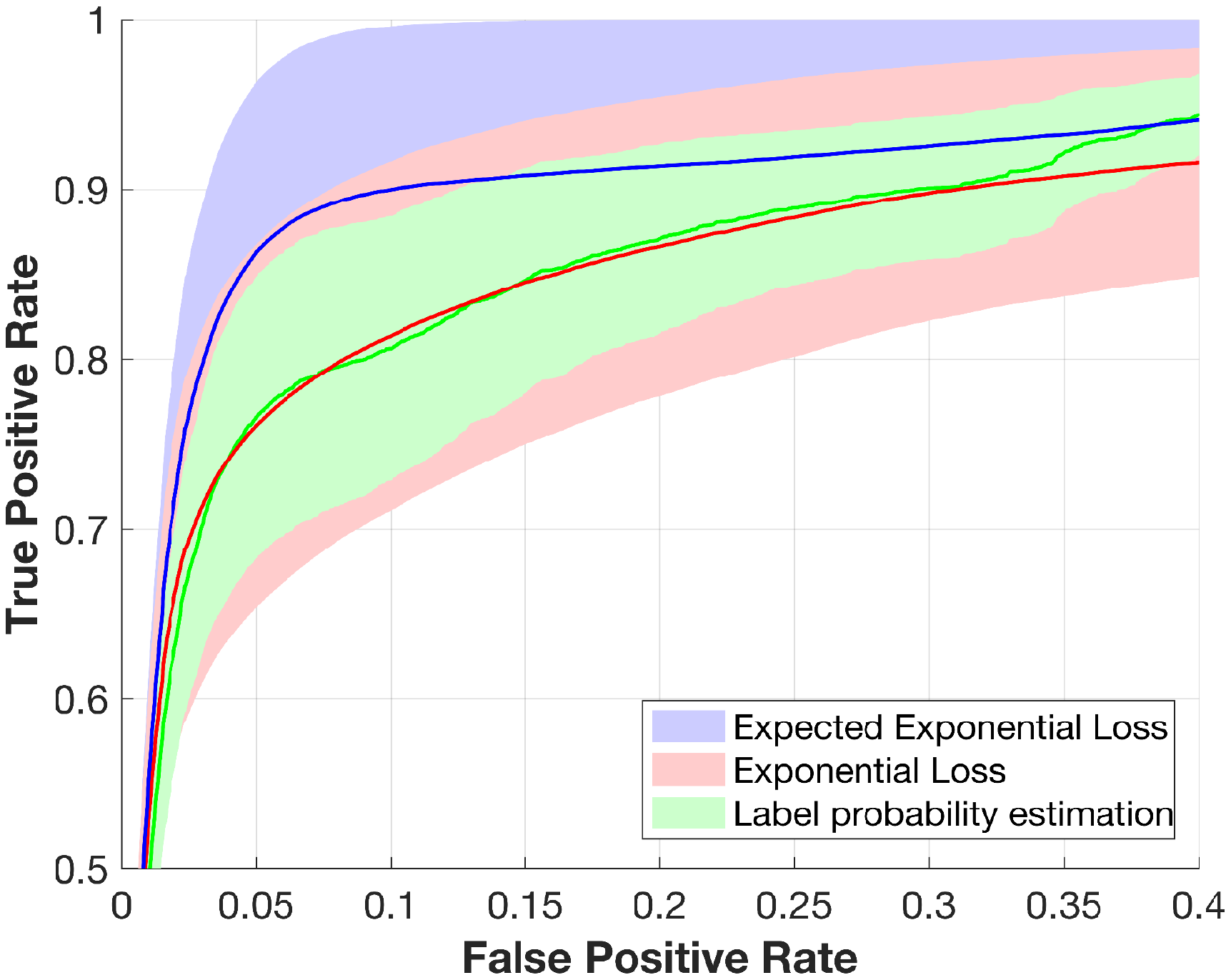}
\includegraphics[width=0.45\textwidth]{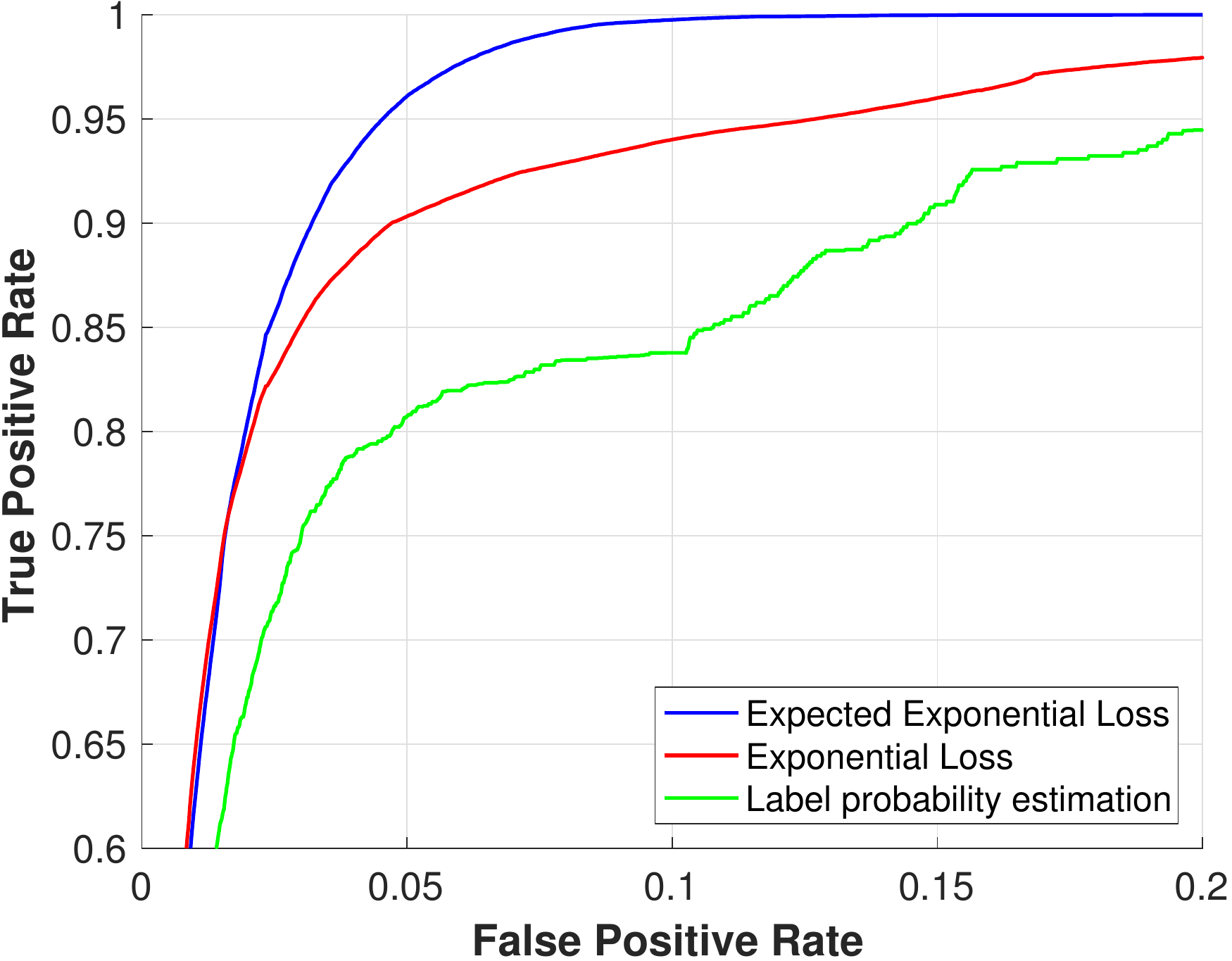}
\caption{(left) ROC performance variability on the same data but induced with different gaze sequences. (right) Performance in a crowd-sourcing setting.}
\label{fig:mutltiesuslt}
\end{figure}

{\bf{Results -- Gaze variance}}: In order to estimate the variance in annotations obtained with our strategy, 7 gaze observations were performed on the same laparoscopic image sequence. From these gaze observations, we ran our method on each set independently. Fig.~\ref{fig:mutltiesuslt}(left) shows the average ROC curve and standard error associated with our approach. In addition, we show similar performances when using the EL and when using the estimated labels only. On average we see that the EL does no better than the label estimation process, while the label estimation approach has slightly less variability. Overall, the EEL approach not only outperforms the other settings, but has lower variance as well. 

{\bf{Results -- Crowd-Sourcing context}}: From the 7 gaze observations collected, we consider a Crowd-Sourcing context where the label estimation is combined as described in Sec.~\ref{sec:probapropa} in order to generate the associated ground truth. Fig.~\ref{fig:mutltiesuslt}(right) illustrates the performance attained when doing so. While the overall trend is no different to the previous experiment, the performance reached by the EEL approach is vastly higher. This is unsurprising given that more gaze information is provided in this setting (\ie 7 annotators) and that more of the object is in fact viewed, yielding thus more positive samples, as well as better $\epsilon_S$ estimates.

\section{Conclusion} 
\label{sec:conclusion}
In this work we have presented a strategy for domain experts to provide useful pixel-wise annotations in a passive way. By leveraging cheap eye gaze tracking technology, we have showed that gaze information can be used to produce segmentation ground truth in a variety of 3D or video imaging modalities. We achieved this by introducing a novel EEL function that is robust to large amounts of unlabeled data and few positive samples. We also demonstrated that our approach could be used in the context of crowd-sourcing where multiple annotators are available.

While this work presents an initial attempt, a number of aspects of this work need to be explored moving forward. In particular, we plan to tackle the case when the object is not present during the entire sequence, as well as cases where multiple objects are present. Naturally, asking more of the user would provide additional information, but our goal is to keep this to a minimum. For this reason, we also plan on determining how our method could work with noisy object observations, as \%100 compliant users may not always be possible. 


\subsubsection*{Acknowledgements:} This work was supported in part by the Swiss National Science Foundation Grant $200021\_162347$ and the University of Bern.

\bibliographystyle{plain}
\bibliography{bibfiles}

\end{document}